\documentclass[journal,10pt]{IEEEtran}
\IEEEoverridecommandlockouts
\usepackage{cite}
 \usepackage{color}

\usepackage{float}
\usepackage{derivative}
\usepackage{amsmath,amssymb,amsfonts}
\usepackage[english]{babel}
\usepackage{amsthm}
\theoremstyle{definition}

\usepackage{graphicx}
\usepackage{caption}
\usepackage{subcaption}
\usepackage{multirow}
\usepackage[left=1.35cm,right=1.35cm,top=1.9cm,bottom=4.1cm]{geometry}
\usepackage{textcomp}
\usepackage{xcolor}
\usepackage{bbm}
\usepackage{algorithm}
\usepackage{algpseudocode}
\newtheorem{theorem}{Theorem}

\def\BibTeX{{\rm B\kern-.05em{\sc i\kern-.025em b}\kern-.08em
    T\kern-.1667em\lower.7ex\hbox{E}\kern-.125emX}}
\usepackage{tikz}
\usepackage[subtle,tracking=normal]{savetrees}
\begin{document}
\def\blue{\textcolor{blue}}
\title{Graph Neural Networks for Resource Allocation in Interference-limited Multi-Channel Wireless Networks with QoS Constraints\\
}
\newcommand\blfootnote[1]{%
  \begingroup
  \renewcommand\thefootnote{}\footnote{#1}%
  \addtocounter{footnote}{-1}%
  \endgroup
}
\author{\IEEEauthorblockN{Lili Chen, Changyang She, Jingge Zhu and Jamie Evans}

\thanks{Part of this work has been submitted to GLOBECOM 2025. L. Chen, J. Zhu and J. Evans are with the Department of Electrical and Electronic Engineering, University of Melbourne, Australia. (Email: \{lili.chen1, jingge.zhu, jse\}@unimelb.edu.au) C. She is with the School of Information Science and Technology, Harbin Institute of Technology (Shenzhen), Shenzhen, China. Part of this work was done when he was with the School of Electrical and Information Engineering, The University of Sydney, Sydney, Australia. (Email: shechangyang@gmail.com)}
}

\maketitle

\begin{abstract}
Meeting minimum data rate constraints is a significant challenge in wireless communication systems, particularly as network complexity grows. Traditional deep learning approaches often address these constraints by incorporating penalty terms into the loss function and tuning hyperparameters empirically. However, this heuristic treatment offers no theoretical convergence guarantees and frequently fails to satisfy QoS requirements in practical scenarios.  Building upon the structure of the WMMSE algorithm, we first extend it to a multi-channel setting with QoS constraints, resulting in the enhanced WMMSE (eWMMSE) algorithm, which is provably convergent to a locally optimal solution when the problem is feasible. To further reduce computational complexity and improve scalability, we develop a GNN-based algorithm, JCPGNN-M, capable of supporting simultaneous multi-channel allocation per user. To overcome the limitations of traditional deep learning methods,  we propose a principled framework that integrates GNN with a Lagrangian-based primal-dual optimization method. 
By training the GNN within the Lagrangian framework, we ensure satisfaction of QoS constraints and convergence to a stationary point. Extensive simulations demonstrate that JCPGNN-M matches the performance of eWMMSE while offering significant gains in inference speed, generalization to larger networks, and robustness under imperfect channel state information. This work presents a scalable and theoretically grounded solution for constrained resource allocation in future wireless networks.
\end{abstract}

\begin{IEEEkeywords}
Joint Resource Management, Graph Neural Networks, Wireless Communication, Quality of Service
\end{IEEEkeywords}

\section{Introduction}\label{sec:Introduction}
As the density of mobile devices increases, interference becomes the bottleneck for improving the data rate in future wireless networks, especially when users have stringent Quality-of-Service (QoS) constraints. To support more devices, future communication systems will use larger bandwidth and more carriers, such as orthogonal frequency division multiple access in the fifth-generation cellular networks (5G) and beyond. Proper allocation of channels and transmit power is pivotal to alleviating the interference in multi-channel wireless networks. Nevertheless, it remains one of the open problems in wireless networks as the complexity of finding the optimal policy grows rapidly as the numbers of mobile devices and channels increase. Motivated by this fact, this work investigates the resource allocation in interference-limited multi-channel wireless networks and aims to design a scalable and low-complexity algorithm that can meet the QoS requirement of each user.
\subsection{Related Works}
Channel and power allocation have been extensively studied in the literature, with a focus on optimizing resource utilization in diverse wireless communication scenarios. The authors of \cite{shi2011iteratively} converted the Weighted Sum Rate Maximization (SRM) problem into a Weighted Mean-Square Error Minimization (WMMSE) problem, achieving convexity for individual users and simplifying local optimization. Similarly,  a Joint Channel and Power Allocation (JCPA) framework for Device-to-Device (D2D) communications was proposed in \cite{feng2013device}, introducing a structured three-step approach. The framework begins with an admission control process to determine the admissible D2D pairs. It then allocates transmission power to these pairs and their potential cellular user (CU) partners. In the final step, a Maximum Weight Bipartite Matching (MWBM) scheme is employed to select the optimal CU partner for each D2D pair to maximize the network sum rate. A similar MWBM algorithm is also demonstrated in \cite{meshgi2015joint} for multicast D2D communication.

It is worth noting that these methods impose restrictive constraints. For example, each D2D pair can access only one subchannel at a time, and each channel can be accessed by no more than one D2D pair, leading to suboptimal spectrum efficiency. To improve the network sum rate, a framework allowing multiple D2D users to share a single subchannel was proposed in \cite{abdallah2018power,yuan2018iterative}. Multi-channel D2D communications have been explored in \cite{elnourani2018underlay}, where each D2D pair is permitted to access multiple channels simultaneously. However, \cite{elnourani2018underlay} imposes a restriction that each channel can only be used by one D2D pair, limiting the spectrum reuse potential. 

To overcome this limitation, the methods in \cite{hajiaghajani2016joint} and \cite{mach2019resource} propose heuristic channel assignment algorithms that allow channels to be shared by multiple D2D pairs, with each D2D pair capable of reusing multiple channels. While these approaches enhance spectral efficiency, they have notable drawbacks. Specifically, \cite{hajiaghajani2016joint} relies on fixed transmission power for CUs, leading to a sub-optimal performance in dynamic network conditions. Additionally, the sequential allocation with interference evaluation employed in this method incurs high computational complexity, scaling cubically with the number of D2D pairs \cite{mach2019resource}, which limits their scalability and practical applicability in large networks.

Despite these advancements, most traditional optimization approaches remain computationally prohibitive for large-scale multi-channel wireless networks. Recent advancements in machine learning have shown promise in addressing the computational challenges of traditional optimization methods. Multi-Layer Perceptrons (MLPs), initially developed for image recognition tasks, have been successfully applied to resource allocation problems \cite{liang2019towards}. For example, the authors of \cite{sun2018learning} employed MLPs for sum-rate maximization in interference channels by mimicking the iterations of the WMMSE algorithm. However, as the network size grows, MLPs suffer from performance degradation, high training complexity, and sensitivity to changes in network conditions \cite{shen2020graph}. Non-stationary wireless environments further exacerbate these limitations, as parameters like channel distributions and traffic patterns drift over time, introducing harmful effects from hidden variables \cite{riley2019three}.

 Graph Neural Networks (GNNs) offer a compelling alternative by leveraging the inherent topology of wireless networks. GNNs update node embeddings using information from neighboring nodes, making them well-suited for communication networks. Compared to MLPs, GNNs demonstrate superior performance in handling graph-structured data, achieving better generalization across unseen network configurations \cite{keriven2019universal}. 
 Studies in \cite{shen2020graph}, \cite{10283511} and \cite{perera2023flex} applied GNNs to power allocation problems for half-duplex, full-duplex and flexible-duplex transmission, respectively, achieving significant improvements in network efficiency and enhancing user experience. GNN-based channel management has also been considered in \cite{gao2022decentralized} and \cite{he2020resource}.

GNNs are also utilized in addressing joint resource allocation challenges. The authors of \cite{chen2021gnn} proposed a GNN to learn the channel allocation given a power allocation scheme. Then, the optimal power allocation is derived based on the channel allocation obtained from the GNN. Nonetheless, their methodology imposes a restriction wherein a channel can be accessed by at most one D2D pair. Different from \cite{chen2021gnn}, the authors of \cite{chen2024gnn} designed a GNN that allows for the concurrent access of a channel by multiple D2D pairs. In \cite{marwani2024graph}, QoS constraints are further considered in their optimization problem, where QoS constraints are the same for all the D2D users. They include QoS constraints as penalty items in the loss function with no theoretical convergence guarantees. However, the proposed algorithm suffers from slow convergence, requiring extensive time to reach a stationary point, which may not be feasible for real-time or dynamic environments. 

\subsection{Motivation and Contributions}
Despite these advancements, the above schemes \cite{chen2021gnn,chen2024gnn,marwani2024graph} restrict each D2D user to access at most one channel. To address the above issues, this paper delves into the design of a GNN framework for the joint optimization of channel and power allocation in multi-channel networks. We will answer the following questions in this paper: 1) How can we design a GNN algorithm that guarantees convergence to a locally optimal solution in the presence of QoS constraints?  2) Furthermore, can such an algorithm overcome the limitations of existing approaches that restrict only one user from being active on a channel at any given time? To illustrate our approach, we consider a traditional optimization problem that maximizes the total sum rate.  The algorithm can be applied to other types of optimization problems, such as maximizing spectrum efficiency. The main contributions of this paper are summarized below, 
\begin{itemize}
    \item  
    We develop an enhanced WMMSE (eWMMSE) algorithm, building upon the iterative framework proposed in \cite{shi2011iteratively}, to address the JCPA problem in multi-channel wireless networks. In this formulation, QoS constraints are incorporated as minimum data rate requirements for each link, ensuring the algorithm's applicability across networks with diverse data rate demands. We prove that the proposed eWMMSE algorithm is guaranteed to converge to a locally optimal solution, provided the problem is feasible.
    \item 
    To reduce computational complexity, we propose a novel Joint Channel and Power allocation GNN algorithm for Multi-channel scenarios (JCPGNN-M). Unlike prior approaches, such as \cite{marwani2024graph}, which limit transceiver pairs to accessing at most one channel, JCPGNN-M enables simultaneous multi-channel allocation and different minimum data rate constraints for each user. This approach aligns with practical requirements, facilitating efficient spectrum reuse and scalable operations in dense wireless networks. 
    \item One key challenge in training GNNs is achieving convergence to a stationary point. To address this, we train the JCPGNN-M using the Lagrangian of the optimization problem as the loss function, where Lagrange multipliers are adjusted during the training process using a primal-dual approach.
    \item 
    We validate the effectiveness of JCPGNN-M in solving sum-rate maximization problems with unlabelled data, eliminating the need for computationally expensive domain-specific knowledge. Extensive simulations demonstrate that JCPGNN-M achieves performance comparable to the eWMMSE algorithm while significantly reducing inference time compared to the convergence time required by iterative optimization algorithms. Moreover, JCPGNN-M consistently outperforms existing solutions in terms of sum rate, QoS satisfaction, and computational efficiency, particularly in large-scale wireless network scenarios.
    \item The proposed algorithms exhibit strong robustness to missing Channel State Information (CSI) and demonstrate excellent generalization capabilities to larger-scale wireless networks. These characteristics make the algorithms highly suitable for real-world applications in diverse and dynamic wireless environments.
\end{itemize}

\section{Resource Allocation problems}\label{sec:system}
In this section, we formulate the joint channel and power allocation problem in a wireless network with multiple channels using different carrier frequencies. 
\subsection{System Model}
\begin{figure}[htbp]
\centerline{\includegraphics[width=9cm]{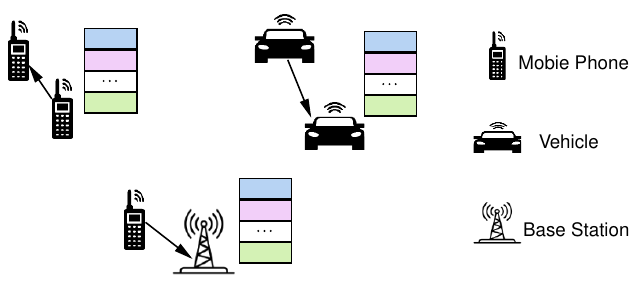}}
\caption{D2D wireless network with multiple channel access.}
\label{C5:fig:systemodel}
\end{figure}

As shown in Fig.~\ref{C5:fig:systemodel}, we consider a wireless network with $D$ transceiver pairs denoted by $\mathcal{D} = \{1,2,...,D\}$, where each transmitter or receiver can be a vehicle, base station, or mobile phone. The transceiver pairs exhibit varying minimum data rate requirements. To meet their requirements, we optimize the channel and power allocation for all transceiver pairs.  Mutual interference arises when two transceiver pairs share the same channel. In Figure~\ref{C5:fig:systemodel}, we employ different colors to indicate different channels. Each transceiver pair can access all the channels at the same time. Let $M$ orthogonal channels, each with identical bandwidth, be at the disposal of the system. We represent the index set for these channels as $\mathcal{M} = \{1, 2, ..., M\}$. In this system model, we follow a similar set-up to \cite{nakashima2020deep} in that we refrain from assigning specific values to the bandwidth of the channels but presume the absence of overlap between them. We assume that each Resource Block (RB) occupies a sufficiently narrow bandwidth such that the channel can be modeled as frequency-flat. Moreover, under the block fading assumption, the CSI varies independently across time slots, requiring independent resource allocation for each frame. The received signal for the $i$-th receiver at $m$-th channel is given by, 
\begin{equation}
    y_{i}^m = h_{i,i}^m s_{i} c_{i}^m+\sum_{j \neq i}h_{i, j}^m s_{j} c_{j}^m+n_{i}^m,
\end{equation}
where $ h_{i,i}^m \in \mathbb{C} $ denote the channel between the $ i $-th transceiver pair over the $ m $-th channel, and $ h_{i,j}^m \in \mathbb{C} $ represent the interference channel from transmitter $ j $ to receiver $ i $ on the same channel. The transmitted signal from transmitter $ i $ is denoted by $ s_i \in \mathbb{C} $. The channel allocation is indicated by the binary variable $ c_i^m $, where $ c_i^m = 1 $ if the $ i $-th transceiver pair utilises the $ m $-th channel for transmission, and $ c_i^m = 0 $ otherwise. The additive Gaussian noise at receiver $ i $ on the $ m $-th channel is modeled as $ n_i^m \sim \mathcal{CN}(0, \sigma_i^2) $. Based on these definitions, the SINR at receiver $ i $ on the $ m $-th channel can be expressed as follows:
\begin{equation}
    \text{SINR}_{i}^{m} = \frac{\left|h_{i,i}^{m}\right|^{2} p_{i}^m c_{i}^{m}}{\sum_{j \neq i}\left|h_{i,j}^{m}\right|^{2} p_{j}^m c_{j}^{m}+\sigma_{i}^{2}},
\end{equation}
where $p_{i}^m$ is the power that the $i$-th transmitter allocates on the $m$-th channel. Denote the channel allocation matrix as $\mathbf{C} = [\mathbf{c}_1,...,\mathbf{c}_D]^{T} \in \mathbb{R}^{D \times M}$, where $\mathbf{c}_i = [c_{i}^1,...,c_{i}^{M}]$ and power allocation matrix as $\mathbf{P} =[\mathbf{p}_1,...,\mathbf{p}_D]^{T} \in \mathbb{R}^{D \times M}$, where $\mathbf{p}_i = [p_{i}^1,...,p_{i}^{M}]$. 
The data rate of the $i$-th transceiver pair over the $m$-th channel is given by,
\begin{equation}
R_{i}^m(\mathbf{C,P})=\log_2 \left(1+\text{SINR}_{i}^{m}\right)\;\text{(bits/s/Hz)}.
\end{equation}
\subsection{Problem Formulation}
The objective of the joint channel and power allocation is to find the optimal channel $\mathbf{C}^{\ast}$ and power profile $\mathbf{P}^{\ast}$ among transceiver pairs at different channels to maximize an objective function under some constraints. In this paper, we focus on the sum rate maximization problem under individual data rate constraints. The problem is formulated as follows, 
\begin{subequations}\label{eq:jointchannelqos}
\begin{align}
    \underset{\mathbf{C,P}}{\operatorname{maximize}} & \sum_{m=1}^{M}  \sum_{i=1}^{D}  \alpha_{i} R_{i}^m(\mathbf{C,P}), \\
        \text { subject to } & p_i^m\geq0, \quad  \forall i \in \mathcal{D},m \in \mathcal{M}, \label{eq:power0}\\
    & c_{i}^m =\{0,1\}, \quad \forall i \in \mathcal{D}, m \in \mathcal{M}\label{eq:jointchannelqos_2nd}\\
    &\sum_{m=1}^{M} c_{i}^m p_{i}^m\leq P_{\max }, \quad \forall i \in \mathcal{D} \label{eq:jointchannelqos_3rd}\\
    & \sum_{m=1}^{M} R_{i}^{m}(\mathbf{C,P}) \geqslant R_i^{\min }, \quad \forall i \in \mathcal{D}\label{eq:jointchannelqos_4th} 
\end{align}
\end{subequations}
where $\alpha_{i}$ is the weight for the $i$-th transceiver pair, $P_{\max }$ denotes the maximum power of the $i$-th transmitter and $R_i^{\min }$ is the minimum data rate required by the $i$-th receiver. Constraint \eqref{eq:power0} ensures that the transmit power of each user is non-negative.  The constraint \eqref{eq:jointchannelqos_2nd} defines the channel allocation variable $c_i^m$, which indicates whether the RB $m$ is allocated to user $i$ (if $c_i^m=1$) or not (if $c_i^m=0$). The constraint \eqref{eq:jointchannelqos_3rd} ensures that the total power allocated to user $i$ across all RBs does not exceed the maximum power $P_{\max}$. The last constraint \eqref{eq:jointchannelqos_4th} ensures that the total sum rare to receiver $i$ across all RBs meets or exceeds the minimum data rate $R_i^{\min }$. Due to interference, $R_{i}^{m}(\mathbf{C,P})$ is non-concave, and hence the feasible region of the problem is non-convex.
\section{Traditional optimization Approach in Resource allocation problems}\label{sec:traditional}
In this section, we extend a traditional optimization approach, WMMSE, to solve the joint channel and power allocation problem with different minimum data rate constraints.  
\subsection{Optimization Simplification}
The original problem involves binary decision variables $ c_i^m \in \{0,1\} $, which introduce combinatorial complexity and make the optimization problem discrete, requiring computationally expensive techniques. Since $ c_i^m $ is binary, the product $ c_i^m p_i^m $ satisfies the following conditions: (1) If $ c_i^m = 1 $, then $ c_i^m p_i^m = p_i^m $, meaning the channel is assigned to the user with a certain power level. (2) If $ c_i^m = 0 $, then $ c_i^m p_i^m = 0 $, indicating that no power is allocated to that channel. Thus, the sum $ \sum_{m=1}^{M} c_i^m p_i^m $ effectively reduces to the summation of $ p_i^m $ over only those channels where $ c_i^m = 1 $.  In Problem \eqref{eq:jointchannelqos}, since users are allowed to access multiple channels simultaneously, we observe that a user is considered active on a channel if and only if $ p_i^m \neq 0 $. Therefore, in the equivalent reformulation, we eliminate $ c_i^m $ and directly treat $ p_i^m $ as the decision variable. By defining $ c_i^m $ in terms of $ p_i^m $ as 

\begin{equation}
c_i^m =
\begin{cases}
1 & \text{if } p_i^m > 0, \\
0 & \text{if } p_i^m = 0.
\end{cases}
\label{eq:c_to_p}
\end{equation}
We reformulate the original problem into the following equivalence problem:  
\begin{subequations}\label{eq:jointchannelqos_sim}
\begin{align}
    \underset{\mathbf{P}}{\operatorname{maximize}} & \sum_{m=1}^{M}  \sum_{i=1}^{D}  \alpha_{i} R_{i}^m(\mathbf{P}), \\
        \text { subject to } & p_i^m\geq0, \quad  \forall i \in \mathcal{D},m \in \mathcal{M},\label{eq:jointchannelqos_sim_1st}\\
    &\sum_{m=1}^{M} p_{i}^m\leq P_{\max}, \quad \forall i \in \mathcal{D} \label{eq:jointchannelqos_sim_2nd} \\
    & \sum_{m=1}^{M} R_{i}^{m}(\mathbf{P}) \geqslant R_i^{\min }, \quad \forall i \in \mathcal{D} \label{eq:jointchannelqos_sim_3rd}
\end{align}
\end{subequations}
We now establish that Problem \eqref{eq:jointchannelqos_sim} is equivalent to the original Problem \eqref{eq:jointchannelqos} in terms of feasibility and optimality. Specifically, we demonstrate that: (a) Every feasible solution of the original problem \eqref{eq:jointchannelqos} corresponds to a feasible solution of the reformulated problem \eqref{eq:jointchannelqos_sim}. (b) The optimal solution of the reformulated problem \eqref{eq:jointchannelqos_sim} is a valid solution of the original problem \eqref{eq:jointchannelqos} and attains the same objective value.  Detailed proof can be found in Appendix \ref{appendix:problem}. To find the optimal solution to problem \eqref{eq:jointchannelqos_sim}, we first define the Lagrangian of the problem as follows,
\begin{align}
&\mathcal{L} (\mathbf{P},\lambda,\mu) \triangleq  -\sum_{m=1}^{M}\sum_{i=1}^{D}\alpha_{i} R_{i}^m(\mathbf{P}) \\
&+\sum_{i=1}^{D} \lambda_{i}\left(\sum_{m=1}^{M} p_{i}^{m}-P_{\text {max }}\right) \nonumber+\sum_{i=1}^{D} \mu_{i}\left(R_i^{\text {min }}-\sum_{m=1}^{M} R_{i}^{m}(\mathbf{P})\right),
\end{align}
where $\lambda_{i}\geq 0$ and $\mu_{i}\geq 0$ are the Lagrange multipliers. We define multiplier vectors $\lambda$ and $\mu$, where $\lambda = [\lambda_1,...,\lambda_D]$ and $\mu = [\mu_1,...,\mu_D]$.  Non-negativity constraint \eqref{eq:jointchannelqos_sim_1st} is handled separately via Karush-Kuhn-Tucker (KKT) conditions \cite{boyd2004convex}. The KKT conditions indicates that the optimal solution should satisfy the following conditions, 
\begin{subequations}
\begin{align}
&\frac{\partial \mathcal{L} (\mathbf{P},\lambda,\mu)}{\partial \mathbf{P} }=0, \label{eq:first-order}\\
& \lambda_{i}\left(\sum_{m=1}^{M}p_{i}^{m}-P_{\text {max }}\right)=0 \quad i \in D, \\
& \mu_{i}\left(R_i^{\text {min}}-\sum_{m=1}^{M} R_{i}^{m}(\mathbf{P})\right) =0 \quad i \in D, \\
& \lambda_{i} \geqslant 0, \mu_{i} \geqslant 0, \quad i \in D,\\
& \eqref{eq:jointchannelqos_sim_1st}, \eqref{eq:jointchannelqos_sim_2nd}, \eqref{eq:jointchannelqos_sim_3rd}.
\end{align}
\end{subequations}
The WMMSE algorithm transforms the SRM problem into a higher-dimensional space by leveraging the well-known MMSE-SINR equality \cite{shi2011iteratively}. Originally, WMMSE was designed for optimization problems where the decision variables are beamforming vectors with complex entries in a single-channel setting. To adapt this approach to our problem, which focuses on power allocation across multiple channels, we modify the algorithm accordingly. Following the approach in \cite{sun2018learning}, we replace $ h_{i,j} $ with $ |h_{i,j}| $ to simplify the implementation. These modifications ensure that the WMMSE framework is applicable to our multi-channel power allocation problem while maintaining computational efficiency.
\subsection{MMSE Equivalence} \label{c5:sub:srm_qos}
To establish the MMSE minimization problem that is equivalent to problem \eqref{eq:jointchannelqos_sim},
we denote $v_{i}^m = \sqrt{p_i^m}$ as the square root of the transmit power the $i$-th transmitter allocated on the $m$-th channel and $e_i^m$ is the corresponding estimated error defined by, 
\begin{equation}
 e_{i}^{m}=\left(u_{i}^{m} |h_{i,i}^{m}| v_{i}^{m}-1\right)^{2}+\sum_{j \neq i}\left(u_{i}^{m} |h_{i, j}^{m}| v_{j}^{m}\right)^{2}+\left(u_{i}^{m}\right)^{2}\left(\sigma_{i}^{m}\right)^{2}, 
    \label{eq:wmmse_e_optimal}
\end{equation}
where $u_{i}^{m}$ is given by \cite{shi2011iteratively},
\begin{equation}
u_{i}^{m}=\frac{|h_{i,i}^{m}| v_{i}^{m}}{\sum_{j=1}^{D}\left|h_{i, j}^{m} v_{j}^{m}\right|^{2}+\left(\sigma_{i}^{m}\right)^{2}}.
    \label{eq:wmmse_u_optimal}
\end{equation}

\begin{theorem}
\label{th:wmmse_qos}
Let $R_{i}^{m}=\log w_{i}^{m}-w_{i}^{m} e_{i}^{m} +1$, the problem \eqref{eq:joint_wmmse_qos}
\begin{subequations}\label{eq:joint_wmmse_qos}
\begin{align}
& \min_{\mathbf{W,U,V}} \sum_{i=1}^{D} \sum_{m=1}^{M} \alpha_{i}(w_i^m e_i^m - log(w_i^m)) \\
& \text { s.t. } \sum_{m=1}^{M} (v_{i}^{m})^2 \leq P_{\max}, \quad i \in D,  \\
& \sum_{m=1}^{M} w_{i}^{m} e_{i}^{m} \leq \sum_{m=1}^{M} \log w_{i}^{m}+M-R_i^{\rm  min}, \quad i \in D, 
\end{align}
\end{subequations}
 is equivalent to the Problem \eqref{eq:jointchannelqos_sim} in the sense that they share the same global optimal solutions for  $v_i^m$. 
\end{theorem}
Here, $w_i^m$ is a positive variable, and the optimization matrices are defined as follows, $\mathbf{W} = [\mathbf{w}_1,...,\mathbf{w}_D]^{T} \in \mathbb{R}^{D \times M}$ where $\mathbf{w}_i = [w_{i}^1,...,w_{i}^{M}]$, $\mathbf{U} =[\mathbf{u}_1,...,\mathbf{u}_D]^{T} \in \mathbb{R}^{D \times M}$ where $\mathbf{u}_i = [u_{i}^1,...,u_{i}^{M}]$ and $\mathbf{V} =[\mathbf{v}_1,...,\mathbf{v}_D]^{T} \in \mathbb{R}^{D \times M}$ where $\mathbf{v}_i = [v_{i}^1,...,v_{i}^{M}]$. 
The proof of Theorem \ref{th:wmmse_qos} is provided in Appendix \ref{appendix:wmmse_qos}. According to Theorem \ref{th:wmmse_qos}, the solution to Problem \eqref{eq:jointchannelqos_sim} can be obtained by solving the equivalent sum-MSE minimization problem formulated in \eqref{eq:joint_wmmse_qos}.  Since the objective function in \eqref{eq:joint_wmmse_qos} is convex with respect to each optimization variable $\mathbf{U}$, $\mathbf{V}$, and $\mathbf{W}$, the problem can be efficiently solved using the block coordinate descent method \cite{shi2011iteratively}. The weight matrix variable $ w_i^m $ is updated as:  
\begin{equation}\label{eq:wupdate}
    w_i^m = (e_i^m)^{-1},
\end{equation}  
where $ e_i^m $ represents the MSE. The update for $ u_i^m $ follows the MMSE solution given in \eqref{eq:wmmse_u_optimal}. Meanwhile, the transmit power update can be decoupled across transmitters,
leading to the following optimization problem:
\begin{equation}
\begin{aligned}
&\mathcal{L}_i\left(\left\{v_{i}^{m}\right\}_{m=1}^{M}, \lambda_{i}, \mu_{i}\right)=\left(1+\mu_{i}\right) \sum_{m=1}^{M}[w_{i}^{m}(|u_{i}^{m} h_{i,i}^{m} v_{i}^{m}-1|^{2}+\\
&\left|u_{i}^{m}\right|^{2}\left(\sigma_{i}^{m}\right)^{2})+\sum_{j \neq i} w_{j}^{m}\left|u_{j}^{m} h_{j, i}^{m} v_{i}^{m}\right|^{2}] -\sum_{i=1}^{D} (\mu_{i}+\alpha_i) \sum_{m=1}^{M} \log w_{i}^{m}\\
&+\sum_{i=1}^{D} \mu_{i}\left(-M+ R_i^{\min }\right) +\lambda_{i}\left(\sum_{m=1}^{M}\left(v_{i}^{m}\right)^{2}-P_{\text {max }}\right).
\label{eq:joint_wmmse_lag_v}
\end{aligned}
\end{equation}
Using the first-order optimality condition in \eqref{eq:first-order}, the square root of the transmission power is given by:
\begin{equation}
v_{i}^{m}(\lambda_{i},\mu_i)=\frac{(1+\mu_i)w_{i}^{m} u_{i}^{m} h_{i,i}^{m}}{(1+\mu_i)w_{i}^{m} |u_{i}^{m} h_{i,i}^{m}|^2+ \sum_{j \neq i} w_{j}^{m}\left|u_{j}^{m} h_{j, i}^{m}\right|^{2}+\lambda_{i}}.
\label{eq:powerwithqos}
\end{equation}
We employ subgradient and bisection methods to update the multipliers $\mu_i$ and $\lambda_i$. Specifically, $\mu_i$ is updated using stochastic gradient ascent (SGA) \cite{sun2023unsupervised}. At the $t$-th iteration, $\mu_i$ is given by:

\begin{align}
&\mu_{i}^{t}=\nonumber \\
&\left[\mu_{i}^{t-1}+\rho_{i}\left(\sum_{m=1}^{M} w_{i}^{m} e_{i}^{m}-\sum_{m=1}^{M} \log w_{i}^{m}-M+R_i^{\min}\right)^{t}\right]^{+}, 
\label{eq:mu_update}
\end{align}
where $\rho_{i}$ is the learning rate of multiplier $\mu_i$. When $\sum_{m=1}^{M} (v_{i}^{m}(0,\mu_i))^2 \leq P_{\max }$, the optimal square root of the transmission power $(v_{i}^{m})_{\text{opt}} = v_{i}^{m}(0,\mu_i)$. Otherwise $v_{i}^{m}(\lambda_{i},\mu_i)$ should satisfy the power constraint,
\begin{equation}
\sum_{m=1}^{M} (v_{i}^{m}(\lambda_{i},\mu_i))^2 \leq P_{\max }. 
\label{eq:powerwithqos_maximum}
\end{equation}

Since $\lambda_i \geq 0$ and $v_{i}^{m}(\lambda_{i},\mu_i)$ is a decreasing function of $\lambda_i$ when $\mu_i$  is fixed, the optimization problem in \eqref{eq:powerwithqos_maximum} can be efficiently solved using the bisection method. Once $\lambda_i$ is determined, we substitute it into \eqref{eq:powerwithqos} to compute $v_{i}^{m}(\lambda_{i},\mu_i)$ . The pseudo-code for the enhanced WMMSE algorithm incorporating QoS constraints is presented in Algorithm \ref{al:wmmse_qos}. If the problem is infeasible, the proposed eWMMSE algorithm will produce a suboptimal power and channel allocation that minimizes QoS constraint violations while maximizing the total achievable rate. Conversely, if the problem is feasible, the eWMMSE algorithm is guaranteed to converge to a locally optimal solution.

\begin{algorithm}
\caption{Pseudo Code of eWMMSE for Joint Channel and Power allocation With QoS Constraints}
\begin{algorithmic}[1]
\State \textbf{Initialization}:Randomly initialize $(v_i^m)^0$ such that  $\sum_{m=1}^{M} (v_{i}^m)^2 \leq P_{\max}, \forall i$; $(\mu_i)^0=0$; $(\lambda_i)^0=0$; set iteration $t=1$.
\State{Calculate } $(u_i^m)^0=\frac{|h_{i, i}^{m}| (v_i^m)^0}{\sum_{j=1}^{D}\left|h_{i, j}^{m} (v_{j}^{m})^0\right|^{2}+\left(\sigma_{i}^{m}\right)^{2}}, \forall i \text {; }$
\State {Calculate } $(w_i^m)^0=\frac{1}{1-(u_i^m)^0\left|h_{i, i}^{m}\right| (v_i^m)^0}, \forall i \text {; }$
\State {Calculate } $ (e_{i}^{m})^0=\left|(u_i^m)^0 h_{i, i}^{m} (v_i^m)^0-1\right|^{2}+\sum_{j \neq i}\left|(u_i^m)^0 h_{i, j}^{m} (v_{j}^{m})^0\right|^{2}+\left|(u_{i}^{m})^0\right|^{2}\left(\sigma_{i}^{m}\right)^{2} \forall i \text {; } $
\While{$t<100$}
\For{$i=1$ to $D$}
\State{Calculate} $\mu_{i}^{t}$ using equation \eqref{eq:mu_update}.
\State{Calculate} $\lambda_i^{t}$ using bisection method such that equation \eqref{eq:powerwithqos_maximum} is satisfied.
\For{$m=1$ to $M$}
\State{Update} $(v_{i}^{m})^t$ using equation \eqref{eq:powerwithqos}.
\State Update $(u_i^m)^t$ using equation \eqref{eq:wmmse_u_optimal}.
\State Update $(w_i^m)^t$ using equation \eqref{eq:wupdate}.
\State Update $ (e_{i}^{m})^{t}$ using equation \eqref{eq:wmmse_e_optimal}.
\EndFor
\EndFor
\EndWhile
\State \textbf{Output} $p_i^m=\left(v_i^m\right)^2, \forall i$.
\end{algorithmic}
\label{al:wmmse_qos}
\end{algorithm}

\section{Graph neural networks for joint channel and power allocation}\label{sec:GNN}
In this section,  we present graph representation for our problem and then propose a GNN-based algorithm that allocates the channels and transmission power to maximize the network sum rate.
\subsection{Learning Approach}
Although an iterative optimization algorithm was introduced in the previous section to solve the problem, its computational complexity remains too high for practical implementation. To address this challenge, we develop a learning-based approach to improve efficiency. In particular, if Problem \eqref{eq:jointchannelqos_sim} is convex and satisfies Slater’s condition, it can be equivalently reformulated as the following optimization problem \cite{boyd2004convex}:
 \begin{equation}
 \begin{aligned}
&\underset{\lambda,\mu} {\operatorname{max}} \hspace{0.1cm}
\underset{\mathbf{P(\mathbf{H, R^{\min}})}} {\operatorname{min}} \hspace{0.2cm}
\mathcal{L} (\mathbf{P(\mathbf{H, R^{\min}})},\lambda,\mu)\\
& \text { s.t. } \lambda_{i} \geqslant 0, \mu_{i} \geqslant 0, \quad i \in D 
\end{aligned}
\label{eq:lagarian}
\end{equation}
where $\mathbf{H} \in \mathbb{R}^{(M \times D)\times(M \times D)}$ is the CSI matrix of all channels and ${\mathbf{R}}^{\min} = [R_1^{\min},...,R_D^{\min}] \in  \mathbb{R}^{1 \times D}$ is the minimum data rate requirements for all the transceivers. The dual variables $\lambda_{i}$ and $\mu_{i}$ need to be non-negative.
Since CSI varies across different channels, we define the matrix $\mathbf{H}$ as follows,
\begin{equation}
\mathbf{H} = \begin{pmatrix}
\mathbf{h_1} & 0 & 0 & \cdots & 0 \\
0 & \mathbf{h_2}& 0 & \cdots & 0 \\
0 & 0 & \mathbf{h_3} & \cdots & 0 \\
\vdots & \vdots & \vdots & \ddots & \vdots \\
0 & 0 & 0 & \cdots & \mathbf{h_M}
\end{pmatrix},
\end{equation}
where $\mathbf{h_m} \in \mathbb{R}^{D \times D}$ for  $m=1,2,...,M$, are the CSI matrix of $m$-th channels. Each $\mathbf{h_m}$ is defined as
\begin{equation}
\mathbf{h_m} = \begin{pmatrix}
h_{1,1}^m & h_{1,2}^m  & \cdots & h_{1,D}^m \\
h_{2,1}^m & h_{2,2}^m & \cdots & h_{2,D}^m \\
\vdots  & \vdots & \ddots & \vdots \\
h_{D,1}^m & h_{D,2}^m  & \cdots & h_{D,D}^m
\end{pmatrix}.
\end{equation}
where $h_{i,i}^m \in \mathbb{C}$ is the CSI of the $i$-th transceiver pair, and $h_{i, j}^m \in \mathbb{C}$ is the interference from transmitter $j$ to receiver $i$ over the $m$-th channel.
Since our problem is non-convex, a local optimal solution of the problem \eqref{eq:jointchannelqos_sim} can be obtained by solving Problem \eqref{eq:lagarian} as established in \cite{luenberger1997optimization}. To determine the solution of the Problem \eqref{eq:lagarian}, we approximate the power allocation policy, $\mathbf{\tilde{P}(\mathbf{H, R^{\min}})}$, by a GNN denoted as $\boldsymbol{G}_p(\omega_p;\mathbf{H, R^{\min}})$, where $\omega_p$ is the trainable parameters of the GNN. 
According to the universal invariant and equivariant graph neural networks theorem\cite{keriven2019universal}, a function defined on graphs can be approximated by a GNN, and the approximation can be uniformly well. 
With the definition in place, problem \eqref{eq:lagarian} can be rewritten as,
\begin{equation}
\begin{aligned}
&\underset{\lambda,\mu} {\operatorname{max}} \hspace{0.1cm}
\underset{\omega_p} {\operatorname{min}} \hspace{0.2cm}
\hat{\mathcal{L}} (\mathbf{\tilde{P}(\mathbf{H, R^{\min}})},\lambda,\mu) =\underset{\lambda,\mu} {\operatorname{max}} \hspace{0.1cm} \underset{\omega_p} {\operatorname{min}} -\sum_{m=1}^{M}\sum_{i=1}^{D}\alpha_{i} \tilde{R}_{i}^m\\
&+ \sum_{i=1}^{D} \lambda_{i}\left(\sum_{m=1}^{M} \tilde{p}_{i}^{m} -P_{\text {max }}\right) +\sum_{i=1}^{D} \mu_{i}\left(R_i^{\text {min }}-\sum_{m=1}^{M} \tilde{R}_{i}^m\right) \\
& \text { s.t. } \lambda_{i} \geqslant 0, \mu_{i} \geqslant 0, \quad i \in D 
\end{aligned}
\label{eq:lagarian_gnn}
\end{equation}
where $\mathbf{\tilde{P}(\mathbf{H, R^{\min}})} = \boldsymbol{G}_p(\omega_p;\mathbf{H, R^{\min}})$ is the power allocation generated by GNN and $\tilde{R}_{i}^m =R_{i}^m(\mathbf{\tilde{P}(\mathbf{H, R^{\min}})})$ is the corresponding data rate. To solve the problem formulated in \eqref{eq:lagarian_gnn}, a primal-dual method can be employed to iteratively update the primal variables $\omega_p$ and the dual variables $\lambda_i$ and $\mu_i$. 

In the $s$-th iteration, the primal variables $\omega_p$ are updated using the Stochastic Gradient Descent (SGD) method, while the dual variables $\lambda_i$ and $\mu_i$ are updated using the SGA method. The update rules are expressed as below, 

\begin{subequations}
\begin{align}
\omega_{p}^{(s+1)} & =\omega_{p}^{(s)}-\phi_{\omega_p} \nabla_{\omega_p} \hat{\mathcal{L}}^{(s)} \\
\lambda_{i}^{(s+1)} & =\left[\lambda_{i}^{(s)}+\phi_{\lambda_{i}}\left(\sum_{m=1}^{M}\tilde{p}_{i}^{m}-P_{\text {max }}\right)\right]^{+}\\
\mu_{i}^{(s+1)} & =\left[\mu_{i}^{(s)}+\phi_{\mu_{i}}\left(R_i^{\text {min }}-\sum_{m=1}^{M} \tilde{R}_{i}^{m}\right)\right]^{+}
\end{align}
\label{eq:multiplier_update}
\end{subequations}
where the operator $[x]^{+} = \text {max}\{0,x\} $ ensures that the dual variables $\lambda_{i} \geqslant 0$ and $ \mu_{i} \geqslant 0$. The learning rates for updating the primal variable $\omega_{p}$, the dual variables $\lambda_{i}$ and $\mu_{i}$ are $\phi_{\omega_p}$, $\phi_{\lambda_{i}}$ and $\phi_{\mu_{i}}$, respectively. The gradient of the Lagrangian $\hat{\mathcal{L}}$ with respect to $\omega_p$ is denoted as $\nabla_{\omega_p} \hat{\mathcal{L}}^{(s)}$.
The GNN is trained by iteratively optimizing $\omega_p$, $\lambda_i$, and $\mu_i$ using the SGD method for the primal variable and the SGA method for the dual variables. This iterative primal-dual update mechanism ensures that the QoS constraints $\sum_{m=1}^{M} \tilde{R}_{i}^{m} \geq R_i^{\text{min}}$ are actively enforced during training. Specifically, the dual variables $\mu_i$ increase when the QoS constraints are violated, thereby amplifying their influence in the Lagrangian. As training progresses, the GNN is guided toward solutions that satisfy both the QoS and power constraints. As demonstrated in \cite{eisen2019learning}, the primal-dual method ensures convergence to a locally optimal solution of the original Problem \eqref{eq:jointchannelqos_sim}. 
\begin{figure*}[htbp]
\centerline{\includegraphics[width=.85\textwidth]{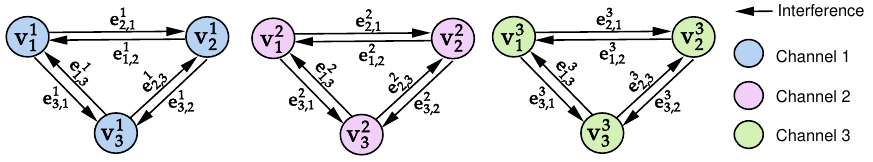}}
\caption{Graph representation of joint resource allocation in multi-channel networks with $D=3$ and $M=3$.}
\label{fig:graphrepre}
\end{figure*}
\subsection{Graph Modeling of Wireless Networks}
A graph $\mathcal{G} = (\mathcal{V}, \mathcal{E})$ represent by a set of vertices $\mathcal{V}$ and a set of edges $\mathcal{E}$. For any two vertices $v_i, v_j \in \mathcal{V}$, an edge $e_{i, j}$ represents a directed connection from $v_j$ to $v_i$. The adjacency matrix of the graph is denoted by $\mathbf{A}$, which is a square matrix where each entry $\mathbf{A}_{(i,j)} \in \{0, 1\}$. Specifically, $\mathbf{A}_{(i,j)} = 1$ if and only if $e_{i, j} \in \mathcal{E}$. 

To effectively capture the CSI within this framework, we model the interference relationships using an interference graph for each channel. Specifically, the $m$-th channel is represented by a subgraph $\mathcal{G}^m$. Since transceiver pairs sharing the same channel interfere with one another, we construct $M$ separate complete graphs, ensuring no loss of information. Figure~\ref{fig:graphrepre} illustrates an example of this graph representation for a system with $D = 3$ transceiver pairs and $M = 3$ channels.

In the subgraph $\mathcal{G}^m$, the $i$-th vertex $v_i^m$ represents the $i$-th transceiver pair transmitting data over the $m$-th channel. Since interference occurs between transceiver pairs sharing the same channel, we define an interference edge $e_{i,j}^m$ between any two vertices $v_i^m$ and $v_j^m$ in the vertex set $\mathcal{V}$. The neighboring set of vertex $v_i^m$ is denoted as $\mathcal{N}(v_i^m) = \{v_j^m \mid e_{i,j}^m \in \mathcal{E}\}$, which includes all vertices connected to $v_i^m$ by interference edges. The node features incorporate properties of the transceiver pairs, such as the direct CSI and the minimum data rate requirements. Specifically, the node feature matrix for vertex $v_i^m$ is defined as $\mathbf{V}_{i}^m = \left[\left|h_{i,i}^m\right|, R_{i}^{\min}\right]$. 
The edge features capture the properties of the interference channels. For each interference edge $e_{i,j}^m$, the edge feature vectors are defined as $\mathbf{E}_{i,j}^m = \left[|h_{i,j}^m|, |h_{j,i}^m|\right]$. 
\subsection{ JCPGNN-M on Graph Neural Networks}
\begin{figure}[htbp]
\centerline{\includegraphics[width=6cm]{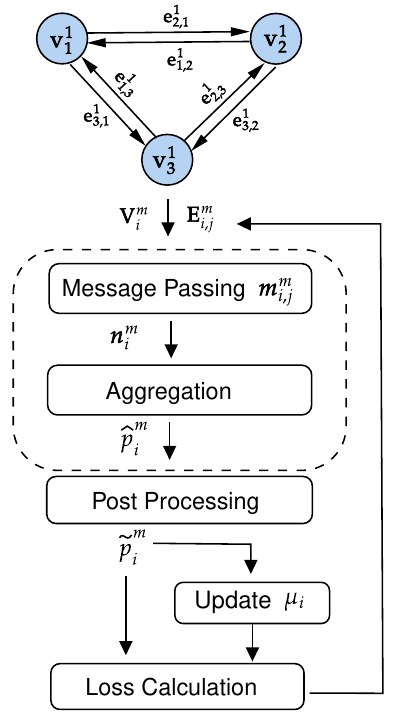}}
\caption{The structure of the proposed JCPGNN-M algorithm.}
\label{fig:GNNstructure}
\end{figure}

By modeling a wireless network as a graph $\mathcal{G}$, the goal is to find a function $\boldsymbol{G}_p(\omega_p;\mathbf{H, R^{\min}})$ mapping each node $v_i^m$ in the $\mathcal{G}$ to the transmit power allocation $\tilde{p}_i^m(\mathbf{H, R^{\min}})$, where $\omega_p$ represents learnable parameters.  Message-Passing Graph Neural Networks (MPGNN), introduced in \cite{shen2020graph}, have been effectively applied to solve radio resource management problems.  
To handle the joint channel and power allocation formulated in problem~\eqref{eq:lagarian_gnn}, we propose the JCPGNN-M algorithm, as illustrated in Figure~\ref{fig:GNNstructure}. The JCPGNN-M architecture consists of three main components: 1) the message-passing layer, 2) the aggregation layer and 3) the post-processing layer.
\subsubsection{Message Computation Layer}
Message computation is a basic unit for GNNs. Since the users have interference when they share the same channel, the message passing will occur within each subgraph $\mathcal{G}^m$. The update rule for message computation in the $s$-th layer for vertex $v_i^m$ is given by,
\begin{subequations}\label{eq:aggreated}
\begin{align}
\boldsymbol{m}_{i,j}^{m (s)}&=\phi_1^{(s)}\left\{\left[\boldsymbol{x}_{i}^{m (s-1)},\mathbf{V}_{i}^m, \mathbf{E}_{i, j}^m\right]: j \in \mathcal{N}(v_i^m)\right\},\\
\boldsymbol{n}_{i}^{m (s)} &=\phi_2^{(s)}\left\{\left[\boldsymbol{m}_{i,j}^{m (s)}\right]\right\},
\end{align}
\end{subequations}
where $\boldsymbol{x}_{i}^{m} = [p_i^m]$ represents the optimization variable (e.g., power allocation) for vertex \(v_i^m\), initialized as \(\boldsymbol{x}_{i}^{m(0)} = 0\). The message computation process consists of two key steps:
First, a nonlinear transformation \(\phi_1^{(s)}(\cdot)\) is applied to generate the message \(\boldsymbol{m}_{i,j}^{m (s)}\). This message combines the vertex's own embedding \(\boldsymbol{x}_{i}^{m (s-1)}\) from the previous layer, its node features \(\mathbf{V}_{i}^m\), and the edge features \(\mathbf{E}_{i, j}^m\) with respect to each neighbor \(v_j^m\). Then, an aggregation function \(\phi_2^{(s)}(\cdot)\) (e.g., SUM) is used to consolidate the messages \(\boldsymbol{m}_{i,j}^{m (s)}\) from all neighbors, producing the aggregated message \(\boldsymbol{n}_{i}^{m (s)}\).

\subsubsection{Aggregation Layer}
The aggregation layer is a fundamental component of GNNs. 
After obtaining the aggregated message \(\boldsymbol{n}_{i}^{m(s)}\) from the message computation layer, the target vertex \(v_i^m\) updates its resource allocation. The update rule for aggregation in the \(s\)-th layer at vertex \(v_i^m\) is defined as:
\begin{equation}
\hat{p}_{i}^{m (s)}=\alpha^{(s)}\left(\boldsymbol{x}_{i}^{m (s-1)}, \boldsymbol{n}_{i}^{m(s)}\right),
\label{eq:gnnaggregate}
\end{equation}
where $\alpha^{(s)}(\cdot)$ is the update functions. It combines the vertex's current state with the aggregated message to produce the updated power allocation \(\hat{p}_{i}^{m (s)}\). To ensure the power allocation values are larger than 0, the output layer of \(\alpha^{(s)}(\cdot)\) employs a Sigmoid activation function.  Typically, MLPs are employed for the aggregation and update functions due to their universal approximation capabilities \cite{hornik1989multilayer}. 

\subsubsection{Post Processing Layer}
As discussed in the previous section, a local optimal solution to problem \eqref{eq:jointchannelqos_sim} can be obtained by solving the Lagrangian formulation in \eqref{eq:lagarian_gnn}. This solution either resides at a stationary point of \(\hat{\mathcal{L}}\) or on the boundary of the feasible region. Consequently, the following optimality conditions must be satisfied:
$\nabla_{\omega_p} \hat{\mathcal{L}} =0$, $\nabla_{\lambda_{i}} \hat{\mathcal{L}}=0$ (or $\lambda_i =0$) and $\nabla_{\mu_i} \hat{\mathcal{L}}=0$ (or $\mu_i =0$). 
Since the maximum power constraint is a strict constraint, we aim to ensure that the solution satisfies \(\nabla_{\lambda_{i}} \hat{\mathcal{L}} = 0\). To achieve this, we employ a normalization technique to adjust the power allocation \(\hat{p}_i^m\). Therefore, we first compute the total power allocation \(\sum_{m=1}^{M} \hat{p}_{i}^{m}\) for each user \(i\). Then, we compare this sum with the maximum power limit \(P_{\max}\). If \(\sum_{m=1}^{M} \hat{p}_{i}^{m} > P_{\max}\), normalize the power allocations \(\hat{p}_{i}^{m}\) such that the total power equals \(P_{\max}\). Otherwise, no normalization is applied. The normalization process is formally defined as:
\begin{equation}
\tilde{p}_{i}^{m}=\hat{p}_{i}^{m} \times \frac{P_{\max}}{\max \left\{\sum_{m=1}^{M} \hat{p}_{i}^{m}, P_{\max}\right\}}
\label{eq:gnn_normalisation}
\end{equation}
\subsection{Loss Function}
Since the proposed JCPGNN-M algorithm is designed to solve Problem \eqref{eq:lagarian_gnn}, its primary objective is to minimize the Lagrangian function. The loss function for JCPGNN-M is formulated as follows:
\begin{equation}
\begin{aligned}
f(\omega_p;\mathbf{H, R^{\min}}) &=-\sum_{m=1}^{M}\sum_{i=1}^{D}\alpha_{i} \tilde{R}_{i}^m+ \sum_{i=1}^{D} \mu_{i}\left(R_i^{\text {min }}-\sum_{m=1}^{M} \tilde{R}_{i}^m\right) \\
&+ \sum_{i=1}^{D} \lambda_{i}\left(\sum_{m=1}^{M} \tilde{p}_{i}^{m}-P_{\text {max }}\right)
\end{aligned}
\end{equation}
The summary of the JCPGNN-M is presented in Algorithm \ref{C5:al:jcpgnn}. Note that if the problem is infeasible, our proposed JCPGNN-M algorithm will generate a suboptimal power and channel allocation that minimizes QoS constraint violations while maximizing the total achievable rate.
\begin{algorithm}
\caption{Pseudo Code of JCPGNN-M Algorithm}\label{alg:cap}
\begin{algorithmic}[1]
\State \textbf{Input}: Graph $\mathcal{G}$; node feature matrix $\mathbf{V}_{i}^m$; edge feature matrix $\mathbf{E}_{i,j}^m$; message transformation $\phi_1^{(s)}(\cdot)$; aggregation function $\phi_2^{(s)}(\cdot)$; update functions $\alpha^{(s)}(\cdot)$;layer number $S$
\State \textbf{Initialization}: Randomly initialize the weight of functions $\phi_1^{(s)}(\cdot)$, $\phi_2^{(s)}(\cdot)$, and $\alpha^{(s)}(\cdot)$; $\boldsymbol{x}_{i}^{m(0)} \gets 0$;
\For{\texttt{$s=1$ to $S$}}
\For{\texttt{$i=1$ to $D$}}
\For{\texttt{$m=1$ to $M$}}
\State Message passing step: Gather the message from neighbors and aggregate them together based on Eq.\eqref{eq:aggreated}.
\State Aggregation step: Calculate the power allocation $\hat{p}_{i}^{m(s)}$ based on Eq.\eqref{eq:gnnaggregate}.
\EndFor
\State Normalize the power allocation $\hat{p}_{i}^{m(s)}$ based on Eq.\eqref{eq:gnn_normalisation}.
\EndFor
 \State Update the multiplier $\mu_{i}^{(s)}$ based on Eq.~\eqref{eq:multiplier_update}.
\EndFor
\end{algorithmic}
\label{C5:al:jcpgnn}
\end{algorithm}

\section{Performance Evaluation}\label{sec:performance_srm}

In this section, we conduct extensive simulations on SRM problems to evaluate the performance of the proposed JCPGNN-M framework. By comparing it with three baseline approaches, we demonstrate the algorithm's generalization capability and robustness to missing CSI.

\subsection{Simulation Settings}
We adopt a system configuration that incorporates both large-scale fading and Rayleigh fading, following the methodologies outlined in \cite{liang2019towards, he2019joint}. The system consists of \(D\) transceiver pairs, where transmitters are spaced 50 meters apart from each other, and each receiver is randomly positioned at a distance between 2 meters and 10 meters from its corresponding transmitter. The CSI between transmitter \(j\) and receiver \(i\) on the \(m\)-th channel is modeled as \(h_{i,j}^m = \sqrt{G_{i,j}} \cdot r_{i,j}^m\), where \(G_{i,j} = \frac{1}{1 + d_{i,j}^{\alpha}}\) represents large-scale fading, \(r_{i,j}^m \sim \mathcal{CN}(0, 1)\) models Rayleigh fading, \(d_{i,j}\) is the distance between transmitter \(j\) and receiver \(i\), and \(\alpha = 2\) is the path loss exponent. The model is trained on 10,000 samples and tested on 1,000 samples to ensure robust evaluation. The hidden layer sizes for the message computation function \(\phi_1(\cdot)\) are \(\{5, 16, 32\}\), and the hidden layer sizes for the update function \(\alpha(\cdot)\) are \(\{33, 16, 8, 1\}\). This configuration ensures a realistic simulation environment, capturing both the deterministic effects of distance-dependent path loss and the stochastic nature of multipath fading. 
\subsection{Benchmark Schemes}
To assess the effectiveness of our proposed guideline, we consider three baselines for performance comparison. The baselines are listed below,
\begin{itemize}
    \item GNN: The GNN algorithm is proposed in \cite{marwani2024graph} for joint channel and power allocation with QoS constraints. To ensure a fair comparison, the networks are trained using unsupervised learning exclusively. In their setup, each user is restricted to accessing only one channel at a time.
    \item ICP-GNN: Independent Channel Power Allocation GNN (ICP-GNN) is designed to independently allocate power to each channel. Each user $i$ on channel $m$ can use a maximum power of $\frac{P_{\max}}{M}$, where $P_{\max}$ is the total available power, and $M$ is the number of channels. This ensures that power is evenly distributed across channels.
    \item Heuristic: We allocate the maximum transmit power $ P_{\max} $ to the channels where user pairs achieve the highest channel gain. For each user pair $ i $, we first identify the channel with the maximum channel gain, denoted as $ |h_{i,i}|_{\max} $, across all available channels. Then, we set $ p_i^m = P_{\max} $ if the channel gain $ |h_{i,i}^m| $ equals $ |h_{i,i}|_{\max} $. Otherwise, $ p_i^m $ is set to 0. 
\end{itemize}
\subsection{Training Convergence Analysis}
\begin{figure}[htbp]
\centerline{\includegraphics[width =.8\linewidth]{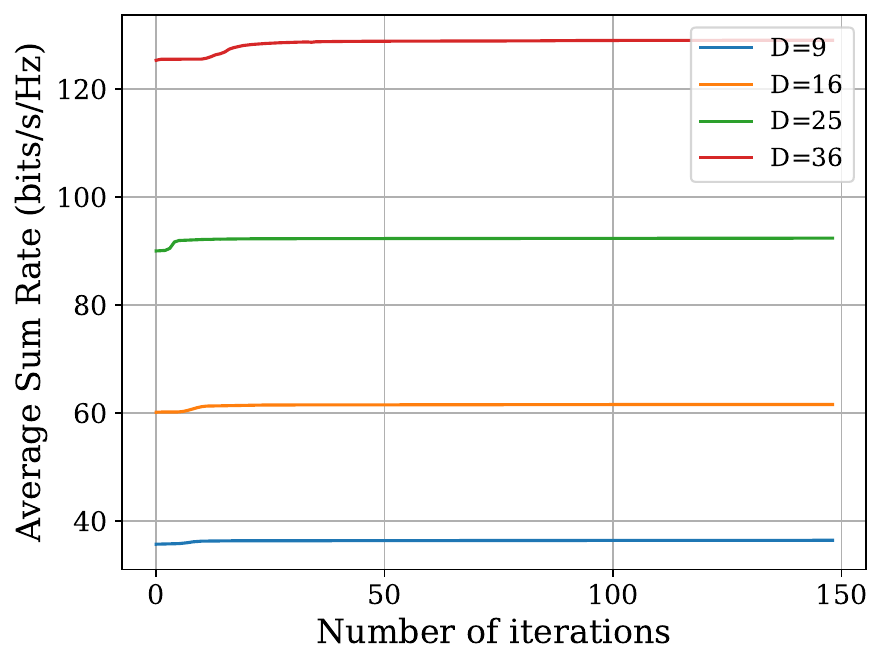}}
\caption{Convergence of average sum rate for different transceiver pairs when $M=2$.}
\label{fig:convergence_allRmin}
\end{figure}

As mentioned in Section~\ref{sec:Introduction}, the GNN can be trained using the primal-dual method to accelerate convergence. Figure~\ref{fig:convergence_allRmin} illustrates the convergence patterns of the average sum rate across different transceiver pairs. The results exhibit a sharp exponential increase in the initial iterations, followed by a gradual stabilization of the sum rate curves as they approach a maximum. Notably, a plateau is observed after approximately the $20$-th iteration, indicating that the model has likely reached a point of diminishing returns in learning.
In comparison to the method proposed in \cite{marwani2024graph}, which requires up to 800 iterations to achieve convergence, our primal-dual approach demonstrates a significant improvement in efficiency, reducing the convergence time by an order of magnitude. This enhanced speed not only accelerates training but also makes the algorithm more practical for real-time applications in dynamic network environments.
\subsection{Performance Comparison}\label{subsec:performance}
\begin{figure}[htbp]
\centerline{\includegraphics[width =.8\linewidth]{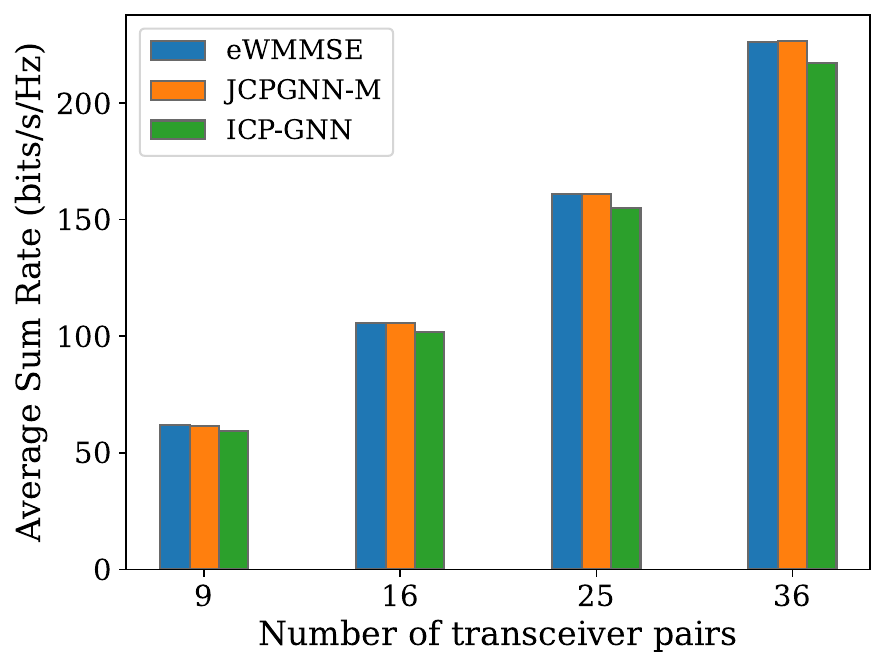}}
\caption{Average sum rate comparison under different minimum data rate requirements for each
transceiver pair when $M=4$.}
\label{fig:sum_DRmin}
\end{figure}

Next, we examine the sum-rate maximization problem under minimum data rate constraints. For each transceiver pair, the minimum data rate constraint is generated from a uniform distribution as $R_i^{\min}\sim\mathcal{U}(1, 2) $. The sum-rate performance for different minimum data rate requirements across transceiver pairs, with $ M = 4 $ channels, is depicted in Figure~\ref{fig:sum_DRmin}. The results demonstrate that the proposed  JCPGNN-M algorithm achieves comparable and, in scenarios with larger $ D $, slightly superior performance to the state-of-the-art eWMMSE algorithm. This enhanced performance can be attributed to  JCPGNN-M's ability to leverage the structural properties of wireless networks, enabling more efficient power allocation, particularly in scenarios involving a large number of channels.
\subsubsection{Impact of QoS Constraints}
\begin{figure}
\centerline{\includegraphics[width =.8\linewidth]{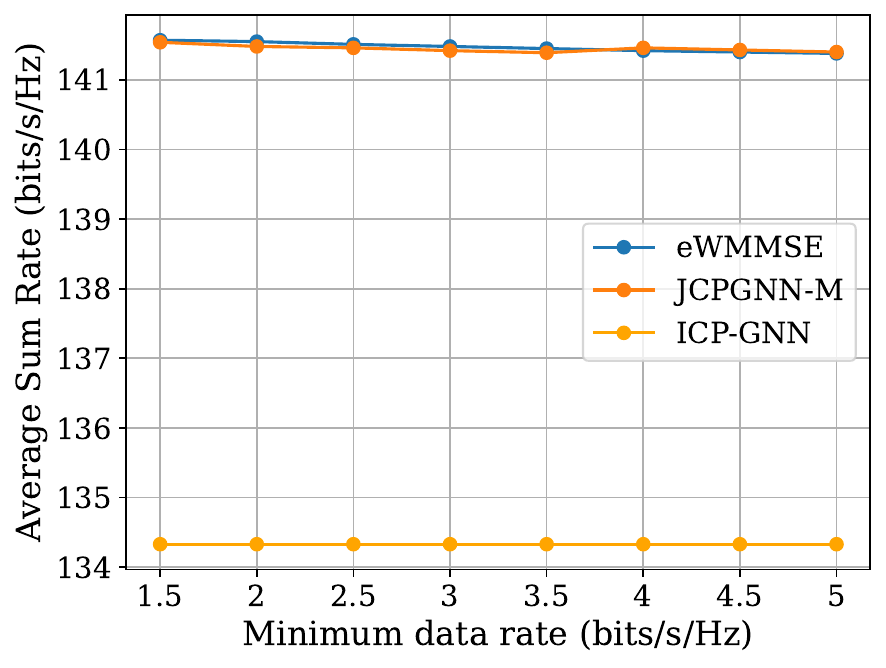}}
\caption{Average Sum rate for different QoS constraints when $M=6$ and $D=16$.}
\label{fig:srm_qos_allRmin}
\end{figure}

\begin{figure}
\centerline{\includegraphics[width =.8\linewidth]{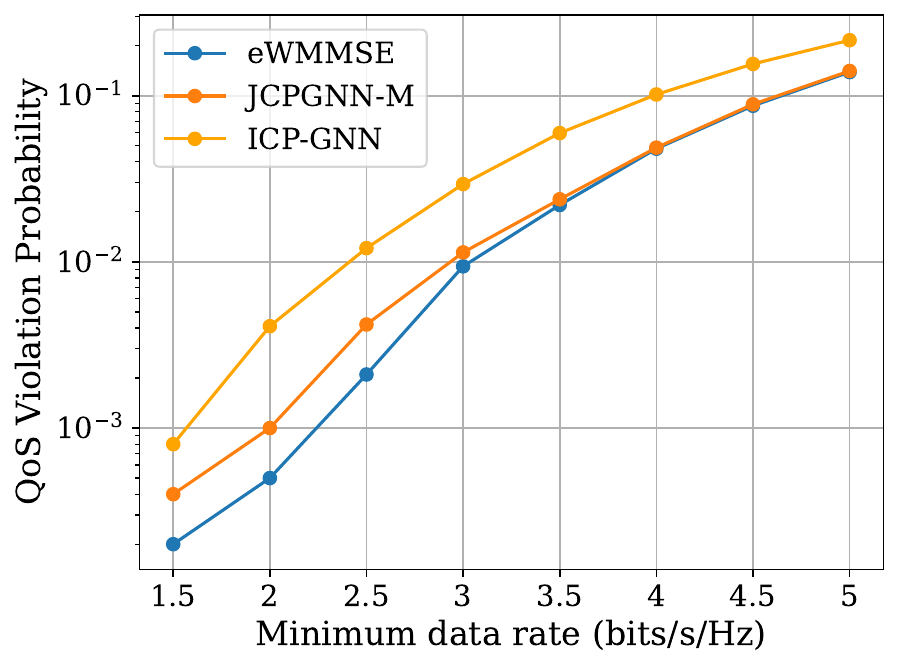}}
\caption{QoS violation probability for different QoS constraints when $M=6$ and $D=16$ (the figure is
presented on a log scale).}
\label{fig:qos_allRmin}
\end{figure}

In this section, we investigate the impact of different minimum data rate constraints. For better comparison, we assume all the transceiver pairs have the same minimum data rate constraint in the following experiments. We fix the number of pairs to $D=16$ and number of channels to $M=6$, and vary $ R_i^{\min}$ from $1.5$ bits/s/Hz to $5$ bits/s/Hz. The performance results of different algorithms are shown in Figure~\ref{fig:srm_qos_allRmin} and Figure~\ref{fig:qos_allRmin}. We observe that the sum-rate performance remains consistent across different minimum data rate constraints. However, the QoS violation probability increases almost exponentially as the minimum data rate increases for all algorithms. eWMMSE and JCPGNN-M exhibit similar performance when \( R_i^{\min} \) is high, demonstrating their robustness in meeting QoS requirements. In contrast, ICP-GNN struggles to satisfy the QoS constraints even at lower minimum data rate values, indicating its limitations in handling stringent QoS conditions. 
As discussed in Section~\ref{c5:sub:srm_qos}, the eWMMSE algorithm is guaranteed to converge to a locally optimal solution if the problem is feasible. Therefore, any residual violation, particularly when $R_i^{\min}$ is large, is likely due to the inherent infeasibility of the problem instance. Our proposed JCPGNN-M maintains a maximum QoS violation gap of less than $2 \times 10^{-3}$ compared to eWMMSE when $R_i^{\min} = 1.5$ bits/s/Hz, and the gap becomes negligible when $R_i^{\min}$ exceeds 3.5 bits/s/Hz. These results confirm that JCPGNN-M can reliably approximate the behaviour of eWMMSE even in borderline-feasible or infeasible regimes. 
\subsubsection{Impact of the Number of Transceiver Pairs and Channels}
\begin{figure*}
\centering
\begin{subfigure}{.32\textwidth}
    \centering
    \includegraphics[width=.95\linewidth]{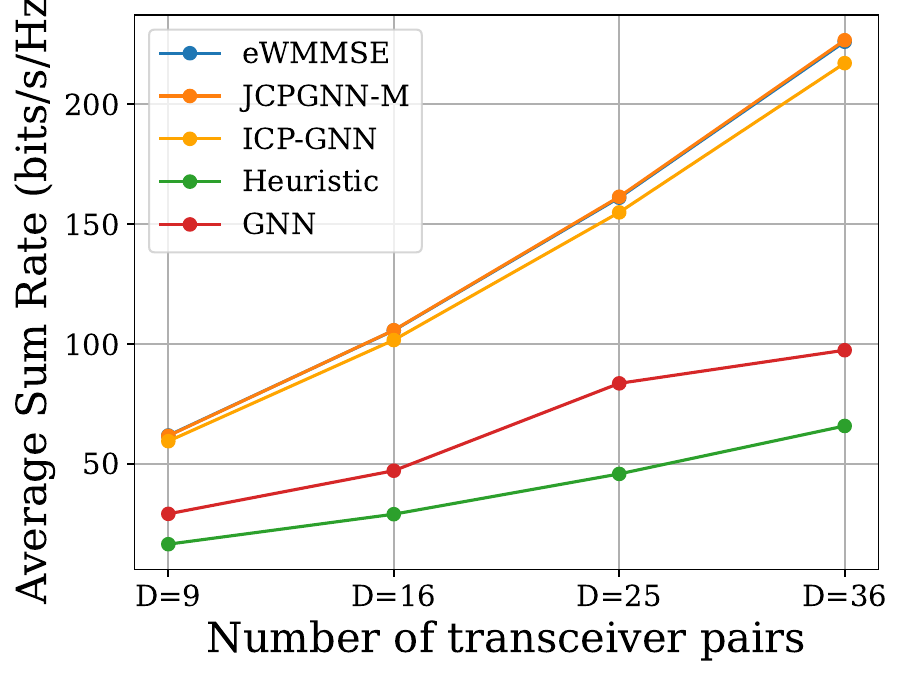}  
    \caption{$M=4$}
\end{subfigure}
\begin{subfigure}{.32\textwidth}
    \centering
    \includegraphics[width=.95\linewidth]{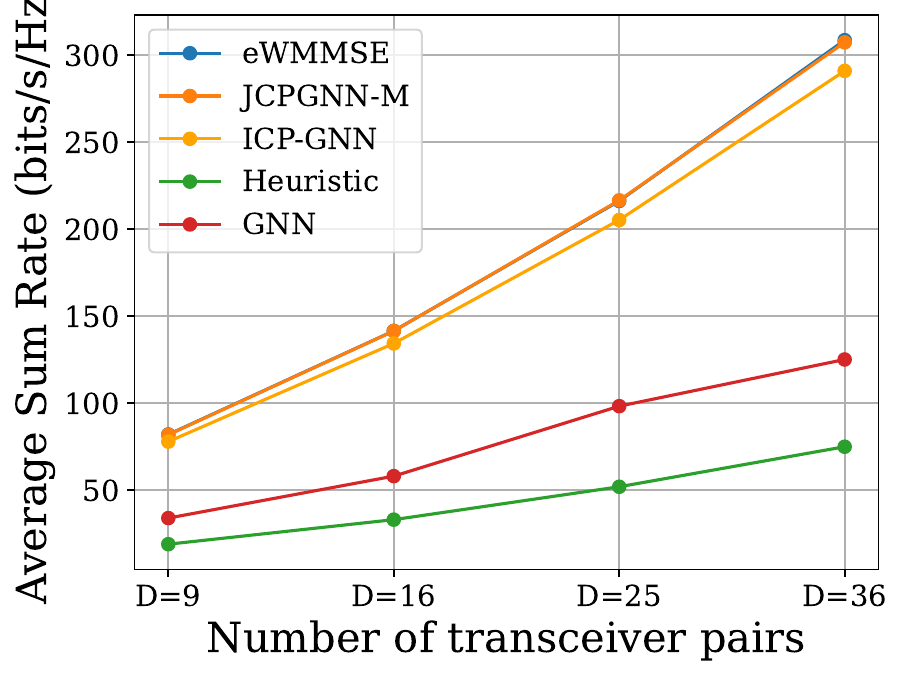}  
    \caption{$M=6$}
\end{subfigure}
\begin{subfigure}{.32\textwidth}
    \centering
    \includegraphics[width=.95\linewidth]{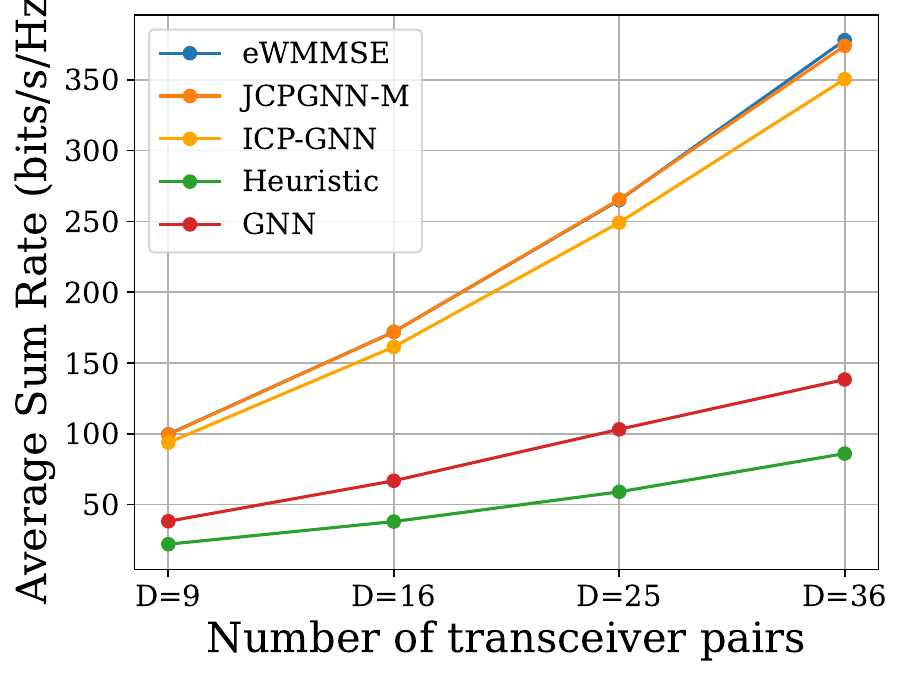}  
    \caption{$M=8$}
    \label{fig:srm_all_5}
\end{subfigure}
\caption{Average sum rate comparison for different number of transceiver pairs when $R_{\min}=2 ~\text{bits/s/Hz}$.}
\label{fig:srm_all_rmin}
\end{figure*}

The performance of different algorithms across varying numbers of channels is shown in Figure~\ref{fig:srm_all_rmin}. In this evaluation, we fix the number of transceiver pairs at \( D = \{9,16,25,36\} \) and set the minimum data rate constraint to \( R_i^{\min} = 2 \) bits/s/Hz, while varying the number of channels as \( M = \{4,6,8\} \).  The results indicate that our proposed JCPGNN-M achieves sum-rate performance comparable to the state-of-the-art eWMMSE algorithm across all scenarios. Moreover, JCPGNN-M outperforms ICP-GNN and GNN, demonstrating its effectiveness in resource allocation. As the number of channels increases while keeping the number of pairs fixed, the performance gap between JCPGNN-M and other methods widens. This is because JCPGNN-M effectively leverages the structure of wireless networks and channel characteristics, enabling more efficient power allocation, particularly in multi-channel environments.

\begin{figure*}
\centering
\begin{subfigure}{.32\textwidth}
    \centering
    \includegraphics[width=.95\linewidth]{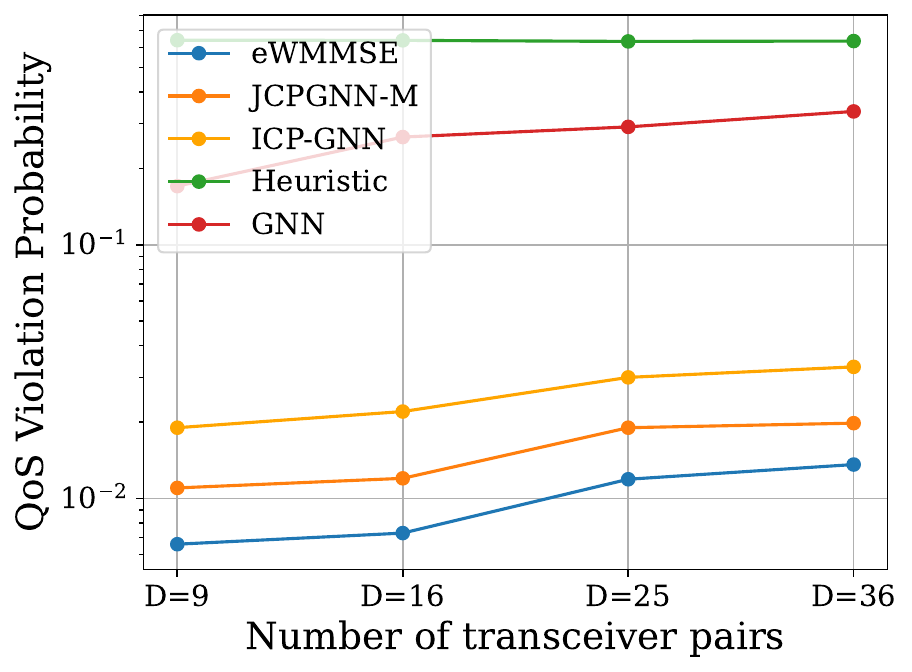}  
    \caption{$M=4$}
    
\end{subfigure}
\begin{subfigure}{.32\textwidth}
    \centering
    \includegraphics[width=.95\linewidth]{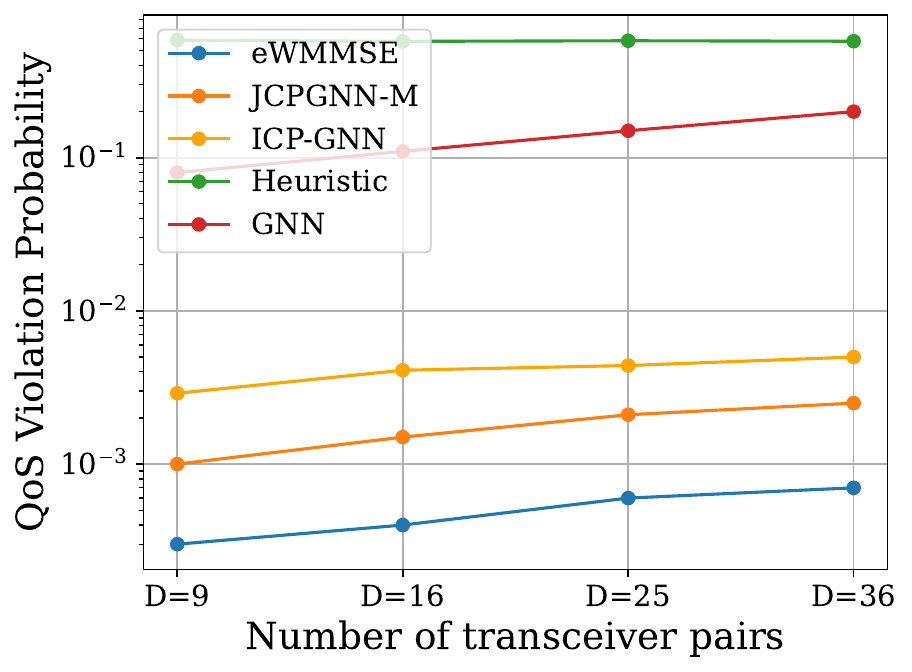}  
    \caption{$M=6$}
    
\end{subfigure}
\begin{subfigure}{.32\textwidth}
    \centering
    \includegraphics[width=.95\linewidth]{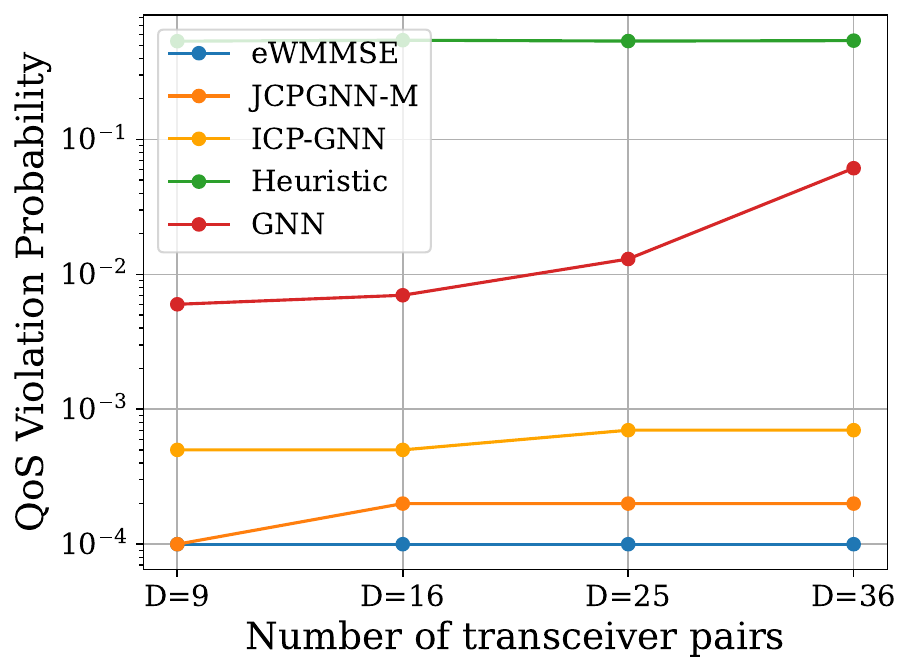}  
    \caption{$M=8$}
\end{subfigure}
\caption{QoS violation probability for different number of transceiver pairs when $R_{\min}=2 \text{bits/s/Hz}$. The y-axis is in log scale.}
\label{fig:qos_all_rmin}
\end{figure*}

The QoS violation probability results of different algorithms for different number of transceiver pairs under the same simulation setup are shown in Figure~\ref{fig:qos_all_rmin}. We can observe the gap between eWMMSE and  JCPGNN-M is very small in all the scenario. The maximum QoS violation probability gap is only $0.7\%$ when $M=4$ and $D=36$. When $M=8$, the gap can be neglect as the violation probability difference is only 4 out of 10000. GNN has the maximum QoS violation probability due to the limitation of exploring the optimal solution with unsupervised learning only.
\subsubsection{Impact of the Missing CSI}
\begin{figure}[htbp]
\centerline{\includegraphics[width =.8\linewidth]{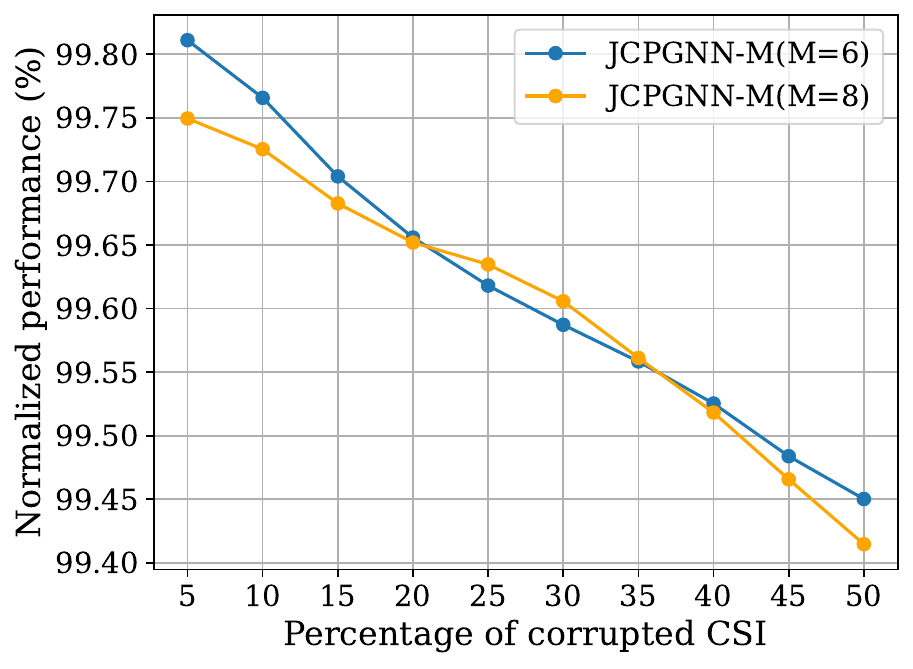}}
\caption{normalized performance for different percentages of missing CSI when $D=16$.}
\label{fig:csi_all}
\end{figure}

Since it is hard to estimate the accurate CSI in dynamic wireless networks, to test the robustness of the proposed  JCPGNN-M algorithm, we train it with missing input CSI. We train the proposed  JCPGNN-M with missing CSI that range from $5\%$ to $50\%$ and test it with the full CSI scenario. The performance of the  JCPGNN-M for $M=6$ and $M=8$ scenarios when $D=16$ is illustrated in Figure~\ref{fig:csi_all}. Note that the performance is normalized by the performance of  JCPGNN-M achieved when training with full CSI. Here, $10\%$ missing CSI input means randomly removing $10\%$ of total CSI when testing the graph neural networks. We can observe from the figure that the proposed  JCPGNN-M can still achieve $99.4\%$ and $99.3\%$ of optimal performance for six and eight-channel scenarios, respectively when $50\%$ of the CSI is missing. This robustness feature is desirable in practical wireless networks where some CSI may be unavailable.
\subsection{Generalization Capacity}\label{sec:gen}
Apart from achieving high sum rate performance, being able to generalize to larger scale problems is also important. In this section, we present the generalization capacity of  JCPGNN-M.
\subsubsection{Generalization to Varied Network Densities}
First, we investigate the generalization capacity to larger network sizes. We trained  JCPGNN-M with smaller networks, e.g., when $D=9$, then tested the algorithm on larger networks, e.g., $D=\{16,25,36\}$. The sum rate performance is shown in Table~\ref{Tab:generalize_qos_users}. The performance is normalized by the maximum sum rate  JCPGNN-M can achieve under the same setup. Our proposed algorithm can achieve the $99\%$ of the best performance even the network sizes is 4 times larger.
\begin{table}[h]
\centering
\caption{normalized sum rate performance for different numbers of transceiver pairs with $M=\{2,4,6,8,10\}$ scenarios.  }
\begin{tabular}{|c|c|c|c|}
\hline \multicolumn{1}{|c|}{ System Scales } & $D=16$ & $D=25$ & $D=36$ \\
\hline{$M=2$}  & $99.7\%$ & $99.5\%$ & $99.1\%$   \\
\hline{$M=4$}   & $99.9\%$ & $99.8\%$ & $99.7\%$ \\
\hline{$M=6$}  & $99.8\%$ & $99.8\%$ & $99.8\%$   \\
\hline{$M=8$}   & $99.8\%$ & $99.8\%$ & $99.9\%$\\
\hline{$M=10$}   & $99.9\%$ & $99.8\%$ & $99.9\%$  \\
\hline
\end{tabular}
\label{Tab:generalize_qos_users}
\end{table}
\begin{table}[h]
\centering
\caption{normalized sum rate performance with different number of channels with $D=\{9,16,25,36\}$ scenarios.}
\begin{tabular}{|c|c|c|c|c|}
\hline \multicolumn{1}{|c|}{ System Scales } & $M=4$ & $M=6$ & $M=8$& $M=10$ \\
\hline{$D=9$}  & $99.7\%$ & $99.1\%$ & $98.6\%$ & $98.3\%$   \\
\hline{$D=16$}   & $99.4\%$ & $99.0\%$ & $98.5\%$ & $98.1\%$ \\
\hline{$D=25$}   & $99.5\%$ & $99.1\%$ & $98.6\%$  & $98.1\%$  \\
\hline{$D=36$}   & $99.7\%$ & $99.4\%$ & $99.2\%$& $98.7\%$ \\
\hline
\end{tabular}
\label{Tab:generalize_qos_channels}
\end{table}
\subsubsection{Generalization to Varied Channel Numbers}
We trained JCPGNN-M using a smaller number of channels (e.g., \( M = 2 \)) and evaluated its generalization performance on larger channel configurations (e.g., \( M = \{4,6,8,10\} \)). The normalized sum rate performance, presented in Table~\ref{Tab:generalize_qos_channels}, demonstrates that our proposed algorithm retains strong generalization capabilities. Notably, JCPGNN-M achieves $98\%$ of the best performance even when tested on channel configurations that are up to five times larger than those used during training.
\subsection{Time Complexity}
\begin{figure}[htbp]
\centerline{\includegraphics[width =.8\linewidth]{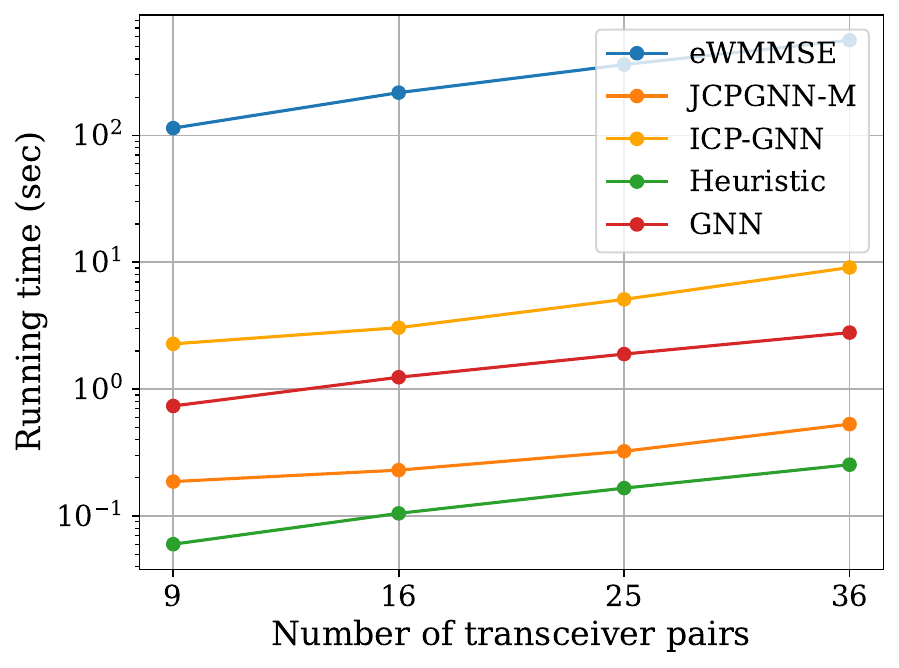}}
\caption{Running time of different algorithms when $M=6$ (the figure is presented on a logarithmic scale).}
\label{fig:time_qos_all}
\end{figure}

The running time of all algorithms for \( M = 6 \) and \( R_i^{\min} = 2 \) bits/s/Hz is presented in Figure~\ref{fig:time_qos_all}, with results shown on a logarithmic scale.  Among the evaluated methods, the heuristic algorithm has the lowest computational complexity due to its simplicity and minimal computational workload. Our proposed JCPGNN-M demonstrates significantly lower complexity compared to both GNN-based benchmarks and the state-of-the-art eWMMSE algorithm. ICP-GNN exhibits higher complexity as it processes each channel individually, increasing its computational burden. Meanwhile, GNN incurs additional overhead due to the use of convolutional neural networks, leading to greater computational costs. As a result, the complexity gap between ICP-GNN and JCPGNN-M widens as the total number of channels increases.   Notably, when \( D = 36 \), our proposed JCPGNN-M is approximately 10 times faster than ICP-GNN and 1,000 times faster than eWMMSE, highlighting its efficiency in large-scale scenarios.

\section{Conclusion}\label{sec:conclusion}
This paper explored a GNN-based joint channel and power allocation framework for multi-channel wireless networks, addressing the limitations of traditional optimization approaches. To solve the problem, we enhanced the WMMSE algorithm from single-channel systems to multi-channel scenarios. By leveraging the block coordinate descent method, we derived optimal power allocation updates using Lagrange multipliers. The reformulation and optimization steps laid the foundation for our proposed JCPGNN-M algorithm, a graph-based learning model designed to efficiently allocate resources in multi-channel environments. Unlike previous methods that restrict users to a single channel at a time, JCPGNN-M enables simultaneous multi-channel access, thereby improving spectral efficiency and overall network performance.

We evaluated the performance of JCPGNN-M under various network conditions, including different numbers of transceiver pairs, channel configurations, and minimum data rate constraints. The results demonstrated that JCPGNN-M generalizes well across different network settings, achieving $98\%$ of the best performance even when applied to scenarios with five times more channels than those used in training. Additionally, our algorithm significantly outperformed existing GNN-based benchmarks and heuristic approaches while maintaining computational efficiency. Compared to eWMMSE, JCPGNN-M achieved comparable or better performance while requiring only $1\%$ of the computational cost, highlighting its scalability for real-time applications.

\appendices 

\section{Proof of the Equivalence of problem \eqref{eq:jointchannelqos} and problem \eqref{eq:jointchannelqos_sim}}\label{appendix:problem}

\subsection{ Proof of Feasibility Preservation}

Claim:  If $\hat{\mathbf{P}}$ is a feasible solution for Problem \eqref{eq:jointchannelqos_sim}, then $(\hat{\mathbf{C}}, \hat{\mathbf{P}})$, where $\hat{\mathbf{C}}$ is determined by Equation~\eqref{eq:c_to_p}, is a feasible solution for Problem \eqref{eq:jointchannelqos}.

Proof: 1. Constraint \eqref{eq:power0} is directly inherited in \eqref{eq:jointchannelqos_sim_1st}. Equation~\eqref{eq:c_to_p} guarantees that constraint \eqref{eq:jointchannelqos_2nd} is satisfied.

2. Power Budget Constraint: Since $\hat{c}_i^m = 1$ if $\hat{p}_i^m>0$ and $\hat{c}_i^m = 0$ if $\hat{p}_i^m = 0$, we substitute this into constraint \eqref{eq:jointchannelqos_3rd} and we have 
\begin{equation}
    \sum_{m=1}^M \hat{c}_i^m \hat{p}_i^m= \sum_{m=1}^M  \hat{p}_i^m\leq P_{\max}
\end{equation}
Thus, constraint \eqref{eq:jointchannelqos_3rd} is satisfied.

3. QoS Constraint: The constraint \eqref{eq:jointchannelqos_4th} can be written as:
\begin{equation}
\sum_{m=1}^M R_i^m(\hat{\mathbf{C}}, \hat{\mathbf{P}}) = \sum_{m=1}^M \hat{c}_i^m R_i^m(\hat{\mathbf{P}}).\end{equation}
Based on the construction of $\hat{c}_i^m$, we have: $\hat{c}_i^m R_i^m(\hat{\mathbf{P}}) = R_i^m(\hat{\mathbf{P}}) $ if $\hat{p}_i^m>0$ and $\hat{c}_i^m R_i^m(\hat{\mathbf{P}}) = 0 $ if $\hat{p}_i^m = 0$, which implies $\sum_{m=1}^M \hat{c}_i^m R_i^m(\hat{\mathbf{P}}) = \sum_{m=1}^M R_i^m(\hat{\mathbf{P}})\geq R_i^{\min}$, where the inequality is guaranteed by \eqref{eq:jointchannelqos_sim_3rd}. Therefore, this constraint is satisfied.

\subsection{Proof of Optimality Preservation}

 We first define $(\mathbf{C}^\star, \mathbf{P}^\star)$ is the optimal solution for Problem \eqref{eq:jointchannelqos}, $\hat{\mathbf{P}}$ is the optimal solution for Problem \eqref{eq:jointchannelqos_sim} and $\hat{\mathbf{C}}$ is determined by Equation~\eqref{eq:c_to_p} based on $\hat{\mathbf{P}}$. To prove the optimality, we aim to show that the optimal solution $(\hat{\mathbf{C}}, \hat{\mathbf{P}})$ of Problem \eqref{eq:jointchannelqos_sim} is also an optimal solution for Problem \eqref{eq:jointchannelqos}. From the definition, the total sum rate is defined as, 
\begin{equation}
\begin{aligned}
    R(\mathbf{C}^\star, \mathbf{P}^\star) &=\sum_{m=1}^M\sum_{i=1}^D \alpha_i R_i^m(\mathbf{C}^\star, \mathbf{P}^\star) = \sum_{m=1}^M\sum_{i=1}^D  \alpha_i (c_i^m)^\star R_i^m(\mathbf{P}^\star) \\
&= \sum_{m=1}^M\sum_{i=1}^D  \alpha_i R_i^m(\mathbf{P}^\star) = R( \mathbf{P}^\star)
\end{aligned}
\end{equation}
Similarly, for Problem \eqref{eq:jointchannelqos_sim}, using the relationship between $\hat{\mathbf{C}}$ and $\hat{\mathbf{P}}$, we have $R(\hat{\mathbf{C}}, \hat{\mathbf{P}}) = R( \hat{\mathbf{P}})$. Assume that $R(\mathbf{C}^\star, \mathbf{P}^\star) > R(\hat{\mathbf{C}}, \hat{\mathbf{P}})$. Substituting the equivalences $R(\mathbf{C}^\star, \mathbf{P}^\star)=R( \mathbf{P}^\star)$ and $R(\hat{\mathbf{C}}, \hat{\mathbf{P}}) = R( \hat{\mathbf{P}})$, this implies 
\begin{equation}
    \sum_{m=1}^M \sum_{i=1}^D \alpha_i R_i^m(\mathbf{P}^\star) > \sum_{m=1}^M \sum_{i=1}^D \alpha_i R_i^m(\hat{\mathbf{P}}).
\end{equation}
However, by definition of the optimization problems:
1:$\hat{\mathbf{P}}$ is the optimal solution for Problem \eqref{eq:jointchannelqos_sim}. 2: $\mathbf{P}^\star$ is a feasible solution for Problem \eqref{eq:jointchannelqos_sim}, as the feasible region of Problem \eqref{eq:jointchannelqos_sim} is a relaxation of Problem \eqref{eq:jointchannelqos}. Thus, by optimality, $ R(\hat{\mathbf{P}}) \geq R(\mathbf{P}^\star)$. This contradicts the assumption $R(\mathbf{P}^\star) > R(\hat{\mathbf{P}})$. Therefore,
the optimal solution $(\hat{\mathbf{C}}, \hat{\mathbf{P}})$ for Problem \eqref{eq:jointchannelqos_sim} is also an optimal solution for Problem \eqref{eq:jointchannelqos}.

\section{Proof of theorem \ref{th:wmmse_qos}} \label{appendix:wmmse_qos}
To see the equivalence, we can check the first optimality condition
to find the optimal $w_i^m$ and $u_i^m$
\begin{equation}
    \begin{aligned}
        & (u_{i}^{m})_{\text{opt}}=\frac{|h_{i, i}^{m}| v_{i}^{m}}{\sum_{j=1}^{D}\left|h_{i, j}^{m} v_{j}^{m}\right|^{2}+\left(\sigma_{i}^{m}\right)^{2}}=|h_{i, i}^{m}| v_{i}^{m} \cdot\left(J_{i}^{m}\right)^{-1}\\
        & (w_i^m)_{\text{opt}} = (e_i^m)^{-1}
    \end{aligned}
\end{equation}
Where $J_{i}^{m} = \sum_{j=1}^{D}\left|h_{i, j}^{m} v_{j}^{m}\right|^{2}+\left(\sigma_{i}^{m}\right)^{2}$. Substituting the  $(u_{i}^{m})_{\text{opt}}$, the optimal MMSE error $e_{i}^{m}$ can be simplified to $(e_{i}^{m})_{\text{opt}} = 1-\left|h_{i, i}^{m}v_{i}^{m}\right|^{2}\left(J_{i}^{m}\right)^{-1}$. Plugging these optimal values in and simplifying \eqref{eq:joint_wmmse_qos} gives the following equivalent optimization problem:
\begin{equation}
    \begin{array}{cl}
    \underset{\mathbf{W,U,V}}{\operatorname{maximize}} & \sum_{m=1}^{M}  \sum_{i=1}^{D}  \alpha_{i}\left(\log \left(\frac{1}{1-\left|h_{i,i}^{m} v_{i}^{m}\right|^{2}\left(J_{i}^{m}\right)^{-1}}\right)\right), \\
    \text { subject to } &\sum_{m=1}^{M} (v_{i}^m)^2 \leq P_{\max }, \quad \forall i \in \mathcal{D},\\
    & \sum_{m=1}^{M} \log \left(\frac{1}{1-\left|h_{i, i}^{m} v_{i}^{m}\right|^{2}\left(J_{i}^{m}\right)^{-1}}\right) \geq R_i^{\min }, \forall i \in \mathcal{D},
    \end{array}
\end{equation}
which is further equivalent to JCPA problem \eqref{eq:jointchannelqos}.

\bibliographystyle{IEEEtran}
\bibliography{main}

\vspace{12pt}

\end{document}